\documentclass[letterpaper, 10 pt, conference]{ieeeconf}  

\IEEEoverridecommandlockouts                              

\usepackage{graphics} 
\usepackage{epsfig} 
\usepackage{mathptmx} 
\usepackage{times} 
\usepackage{amsmath} 
\usepackage{amssymb}  
\usepackage{algorithm}
\usepackage{algorithmic}
\usepackage{multirow}
\usepackage{cite}
\usepackage{subfigure}
\usepackage{stfloats}
\usepackage{capt-of}
\makeatletter
\let\NAT@parse\undefined
\makeatother
\usepackage{url}
\usepackage[table]{xcolor}
\usepackage{colortbl}
\usepackage{hyperref}
\hypersetup{colorlinks=true, linkcolor=red, urlcolor=blue, citecolor=green, anchorcolor=blue}

\makeatletter
\newcommand{\removelatexerror}{\let\@latex@error\@gobble}
\makeatother

\title{\LARGE \bf
Efficient Learning of A Unified Policy For Whole-body Manipulation and Locomotion Skills
}

\author{Dianyong Hou$^{1}$, Chengrui Zhu$^{1}$, Zhen Zhang$^{1}$, Zhibin Li$^{3}$, Chuang Guo$^{1}$ and Yong Liu$^{2, \dag}$ 
\thanks{*This work was supported by the Key R\&D Project of Zhejiang Province under Grant 2024C01172}
\thanks{$^{1}$Institute of Cyber-Systems and Control, Zhejiang University, China.}%
\thanks{$^{2}$State Key Laboratory of Industrial Control Technology of Zhejiang University, China.}%
\thanks{$^{3}$Department of Computer Science, University College London, London, United Kingdom.}%
\thanks{ $^\dag$Corresponding authors.}
}
\begin{document}

\let\oldtwocolumn\twocolumn
\renewcommand\twocolumn[1][]{%
    \oldtwocolumn[{#1}{
    \begin{center}
       \centering
    \begin{minipage}{\textwidth}
        \centering
        \includegraphics[width=0.97\textwidth]{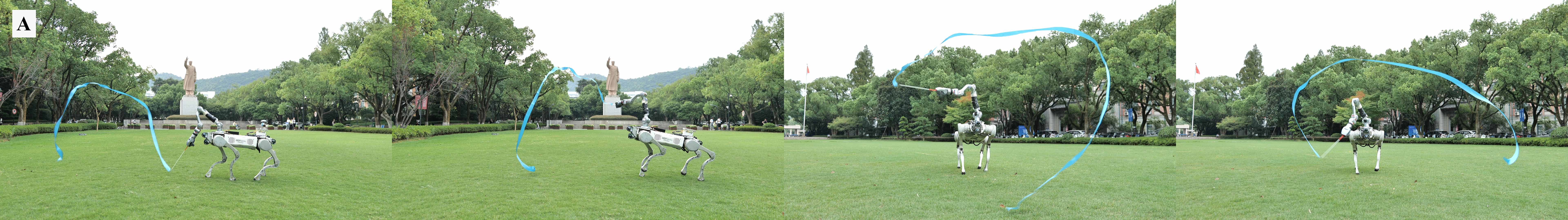}
    \end{minipage}
    \begin{minipage}{\textwidth}
        \centering
        \includegraphics[width=0.97\textwidth]{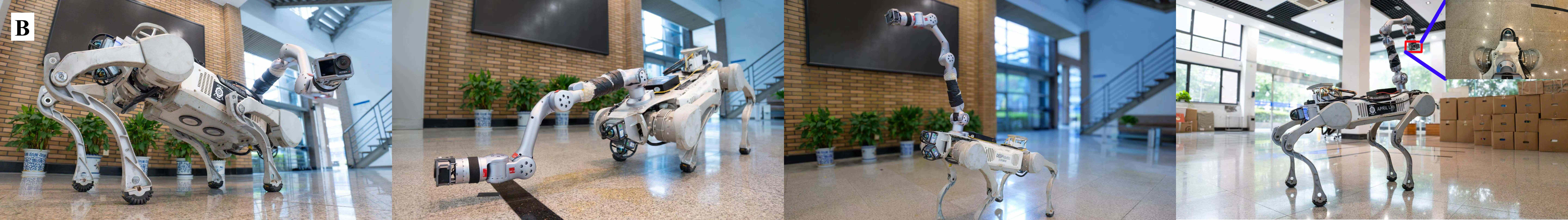}
    \end{minipage}
    \begin{minipage}{\textwidth}
        \centering
        \includegraphics[width=0.97\textwidth]{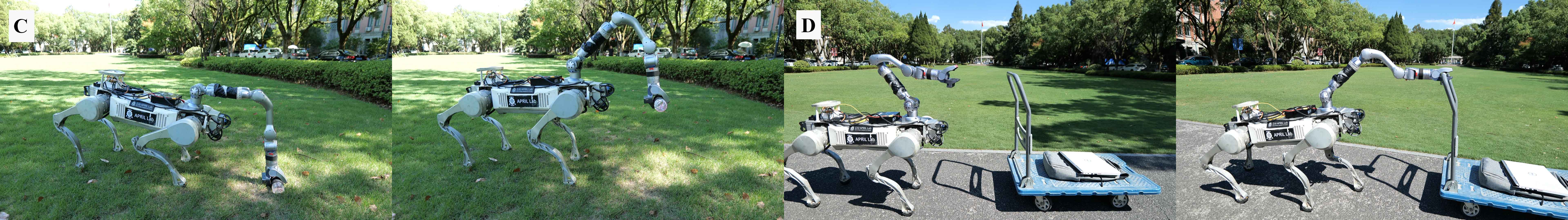}
    \end{minipage}
    \vspace{-3mm}
    \captionof{figure}{A unified learning framework that integrates the kinematic model to effectively guide reinforcement learning, enabling precise whole-body motor skills. (A) demonstrates the ability to mimic the human gymnastic motion of waving a 4-meter ribbon, showcasing highly dynamic and large-range motion control.
    (B) depicts the robot's coordinated movements during a filming task, showcasing its whole-body synergistic control and wide-area reaching capabilities.
    (C) exhibits precise whole-body manipulation by naturally kneeling its front legs to grasp objects from the ground, emphasizing its dexterity and adaptability.
    (D) validates its capability of cart-pushing tasks with payloads, demonstrating its practical applicability in real-world scenarios.
    }\label{fig:whole-body skills}
    \end{center}
    }]
}
\maketitle
\thispagestyle{empty}
\pagestyle{empty}
\begin{abstract}
Equipping quadruped robots with manipulators provides unique loco-manipulation capabilities, enabling diverse practical applications. This integration creates a more complex system that has increased difficulties in modeling and control. Reinforcement learning (RL) offers a promising solution to address these challenges by learning optimal control policies through interaction. Nevertheless, RL methods often struggle with local optima when exploring large solution spaces for motion and manipulation tasks. To overcome these limitations, we propose a novel approach that integrates an explicit kinematic model of the manipulator into the RL framework. This integration provides feedback on the mapping of the body postures to the manipulator's workspace, guiding the RL exploration process and effectively mitigating the local optima issue. Our algorithm has been successfully deployed on a DeepRobotics X20 quadruped robot equipped with a Unitree Z1 manipulator, and extensive experimental results demonstrate the superior performance of this approach. We have established a \href{https://ecstayalive.github.io/posts/PhysicsFeasibilityGuidedOptimization/}{project website} to showcase our experiments. 
\end{abstract}
\section{INTRODUCTION}
Quadruped robots significantly surpass traditional wheeled robots in their ability to adapt to diverse terrains. The integration of a manipulator further enhances their functionality, expanding their application scenarios and enabling them to either substitute or assist humans in performing complex operational tasks.
However, the combination of manipulator and quadruped robots forms a high-degree-of-freedom nonlinear control system, making whole-body cooperative control a particularly challenging endeavor. Model predictive control (MPC) has emerged as a promising solution for addressing whole-body control in legged manipulators. By formulating a comprehensive whole-body dynamics model \cite{bellicoso2019alma, sleiman2021unified, sleiman2023versatile}, this model-based approach enables precise control of both the manipulator and the body. 
Despite its effectiveness, the implementation of MPC demands considerable engineering effort, as it involves the design and integration of multiple interconnected modules. Moreover, to ensure the success of numerical methods in tracking the end-effector pose of the manipulator, it is essential to avoid singularities, which can compromise control accuracy and stability.

In contrast, deep reinforcement learning (DRL) eliminates the need for intricate modeling procedures through its model-free approach while maintaining robustness. It has demonstrated performance comparable to MPC methods in tasks such as velocity tracking \cite{hwangbo2019learning, lee2020learning, miki2022learning, jenelten2024dtc} and dexterous manipulation \cite{gu2017deep, kalashnikov2018scalable, chen2022system, chen2023visual}. Nevertheless, applying DRL to legged manipulators remains challenging due to the interdependent and occasionally conflicting objectives of the arm and body. For instance, to ensure effective manipulation, the quadruped's body must perform cooperative motions, which can compromise locomotion performance. Recent research proposes various solutions to address this issue. Existing works \cite{fu2023deep,portela2024learning,wang2024arm} have shown the effectiveness of RL in jointly controlling quadruped robot velocity and manipulator end-effector position. However, these approaches are susceptible to local optima\cite{fu2023deep}, limiting their effectiveness.

In this paper, we introduce an explicit kinematic model of the legged manipulator to establish a mapping relationship between the body posture and the manipulator's workspace. By incorporating a reward based on this mapping, our algorithm effectively guides the learning process, preventing convergence to local optima. This physical feasibility-guided(PFG) reward also promotes the exploration of whole-body coordinated motor skills, seamlessly integrating the movements of the legs and the manipulator. Our approach has been successfully implemented on a DeepRobotics X20 robot equipped with a Unitree Z1 manipulator. Experimental results demonstrate that our method significantly enhances the whole-body cooperative behavior of the legged robot, resulting in naturally coordinated and smooth movements.

In summary, our contributions are as follows:
\begin{itemize}
\item We incorporate the explicit kinematic model into the RL training framework, resulting in higher execution accuracy compared to other DRL methods.
\item We introduce a physical feasibility-guided reward utilizing kinematics information, effectively preventing RL algorithm convergence to local optima. 
\item  We demonstrate the effectiveness of our method via extensive experiments, showing the unlocked potential applications of legged manipulators in real-world tasks.
\end{itemize}

\section{RELATED WORKS}
\begin{figure}[!t]
    \centering
    \includegraphics[width=\linewidth]{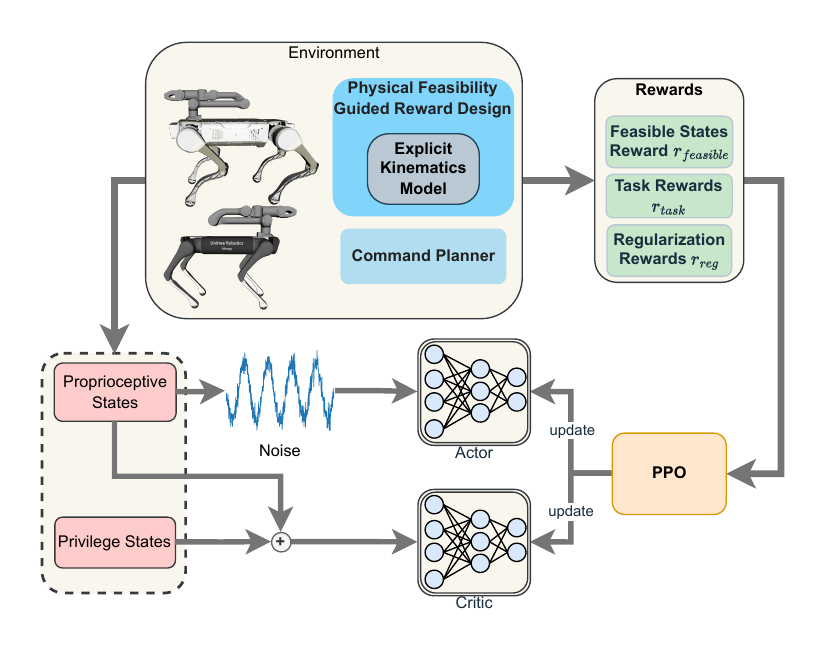}
    \vspace{-5mm}
    \caption{\textbf{Reinforcement learning framework for whole-body loco-manipulation}.} 
    \label{arch}
    \vspace{-5mm}
\end{figure}
\subsection{MPC For Whole Body Control}
MPC methods for whole-body control of legged manipulators require the development of complex dynamics and kinematics models \cite{rehman2016towards, li2022whole, humphreys2022teleoperating}. \textit{Bellicoso et al.} \cite{bellicoso2019alma} introduced the ALMA platform, showcasing impressive locomotion and manipulation capabilities through a unified dynamics model that includes both the quadruped robot and the manipulator. ALMA employs the Zero-Moment Point (ZMP) method to optimize execution trajectories and uses inverse dynamics for whole-body control, achieving coordinated movements between the body and the arm through computational methods. Subsequent research, such as \cite{sleiman2021unified}, expanded upon the whole-body dynamics model and incorporated a broader range of constraints, enabling the controller to handle tasks like door opening and box pushing. \textit{Sleiman et al.} \cite{sleiman2023versatile} further optimized the methods, allowing the body and arm to exhibit excellent coordinated movement capabilities, even granting the legs of the quadruped robot some manipulation abilities. However, the application of MPC for whole-body control remains challenging due to the need for meticulous module design and parameter tuning. For the manipulator, trajectory planning is also necessary to avoid numerical instabilities near singularities in the algorithm.

\subsection{RL For Legged Manipulator}
DRL can avoid the tedious modeling procedures required in traditional control methods. However, training a unified policy for legged manipulators to achieve optimal performance is challenging. Consequently, independent control is often adopted for the body and arm. For example, \cite{ma2022combining} employs an MPC-based arm controller alongside a deep RL-based motion controller. In their approach, the motion of the manipulator is treated as a disturbance to the quadruped robot's RL controller. While this method enables each component to perform its respective task effectively, the lack of cooperation between the separate control mechanisms means the two systems could only passively adapt to disturbances introduced by each other. To overcome these limitations, \textit{Fu et al.} \cite{fu2023deep} introduced an advantage mixing to simplify the training process and successfully developed a whole-body controller. Their work focuses on extending the reachability of the manipulator and makes significant progress. Building on this foundation, \textit{Portela1 et al.} \cite{portela2024learning} presented a force-controlled whole-body controller. 

\section{METHOD}
In this paper, we employ the Proximal Policy Optimization (PPO)\cite{schulman2017proximal} algorithm to address the whole-body cooperative control problem of a legged manipulator. The whole-body control problem is decomposed into two subproblems: velocity tracking control for the quadruped robot and 6D pose tracking control for the manipulator's end-effector. To effectively learn the dynamic coupling between the body motion and the manipulator's operation, we integrate an explicit kinematic model into the RL training framework and introduce a reward design method to guide the policy to learn smooth whole-body control. This reward design method is referred to as the Physical Feasibility-Guided (PFG) reward design method. Additionally, we design a command planning module based on cubic polynomial interpolation to generate reference trajectories for the end-effector and this approach reduces the computational burden associated with time-consuming inverse kinematics iterations. Furthermore, to bridge the sim-to-real gap, we employ an asymmetric actor-critic architecture \cite{pinto2018asymmetric} and adopt domain randomization techniques, including variations in mass and friction coefficients. The overall framework is illustrated in  Fig. \ref{arch}.

\subsection{Observation and Action Space} \label{asymmetric_actor-critic_section}
\begin{table}[!t]
\caption{Observations}
\label{observation_table}
\vspace{-5mm}
\begin{center}
\begin{tabular}{c|c}
\hline
 \multirow{7}*{Actor} & command(body velocity and target pose in body frame) $\in \mathbb{R}^9$ \\
 ~ & quadruped body's roll and pitch angles $\in \mathbb{R}^2$ \\
 ~ & noisy joint position $\in \mathbb{R}^{18}$ \\
 ~ & noisy joint velocity $\in \mathbb{R}^{18}$ \\
 ~ & noisy quadruped body's twist $\in \mathbb{R}^{6}$ \\
 ~ & noisy end-effector pose under body frame $\in \mathbb{R}^{6}$ \\
 ~ & last output action $\in \mathbb{R}^{18}$ \\
 \hline
 \multirow{18}*{Critic} & command(body velocity and target pose in body frame) $\in \mathbb{R}^9$ \\
 ~ & joint states(position, velocity, torque, torque rate) $\in \mathbb{R}^{72}$\\
 ~ & the planned end-effector twist $\in \mathbb{R}^6$ \\
 ~ & the desired manipulator joint position $\in \mathbb{R}^6$  \\
 ~ & inverse kinematics solver status $\in \mathbb{R}^1$ \\
 ~ & quadruped body twist $\in \mathbb{R}^6$ \\
 ~ & action history $\in \mathbb{R}^{36}$ \\
 ~ & foot clearance $\in \mathbb{R}^4$ \\
 ~ & foot period $\in \mathbb{R}^4$ \\
 ~ & foot position $\in \mathbb{R}^{12}$ \\
 ~ & foot slip $\in \mathbb{R}^{4}$ \\
 ~ & foot touchdown impulse $\in \mathbb{R}^{4}$ \\
 ~ & torso disturbance $\in \mathbb{R}^6$ \\
 ~ & end-effector pose under body frame $\in \mathbb{R}^6$  \\
 ~ & end-effector twist $\in \mathbb{R}^6$ \\
 ~ & end-effector disturbance $\in \mathbb{R}^6$ \\
 ~ & manipulator links' mass $\in \mathbb{R}^6$ \\
 ~ & manipulator's jacobian matrix $\in \mathbb{R}^{36}$ \\
 \hline
\end{tabular}
\end{center}
\end{table}
We employ an asymmetric actor-critic architecture. The actor, deployed on the physical robot, receives observations subject to hysteresis and random Gaussian noise, enhancing its robustness. It directly outputs the legged manipulator's $18$ target joint positions. The critic, obtaining information directly from the simulation environment, observes a comprehensive set of data to ensure accurate prediction. This structure has proven to be effective in improving policy performance\cite{ma2023learning}. TABLE \ref{observation_table} lists the detailed observations. 

\subsection{Physical Feasibility-Guided(PFG) Reward Design}
\begin{figure}[!t]
    \centering
    \subfigure{
        \centering
        \begin{minipage}{\linewidth}
            \includegraphics[height=2.76cm]{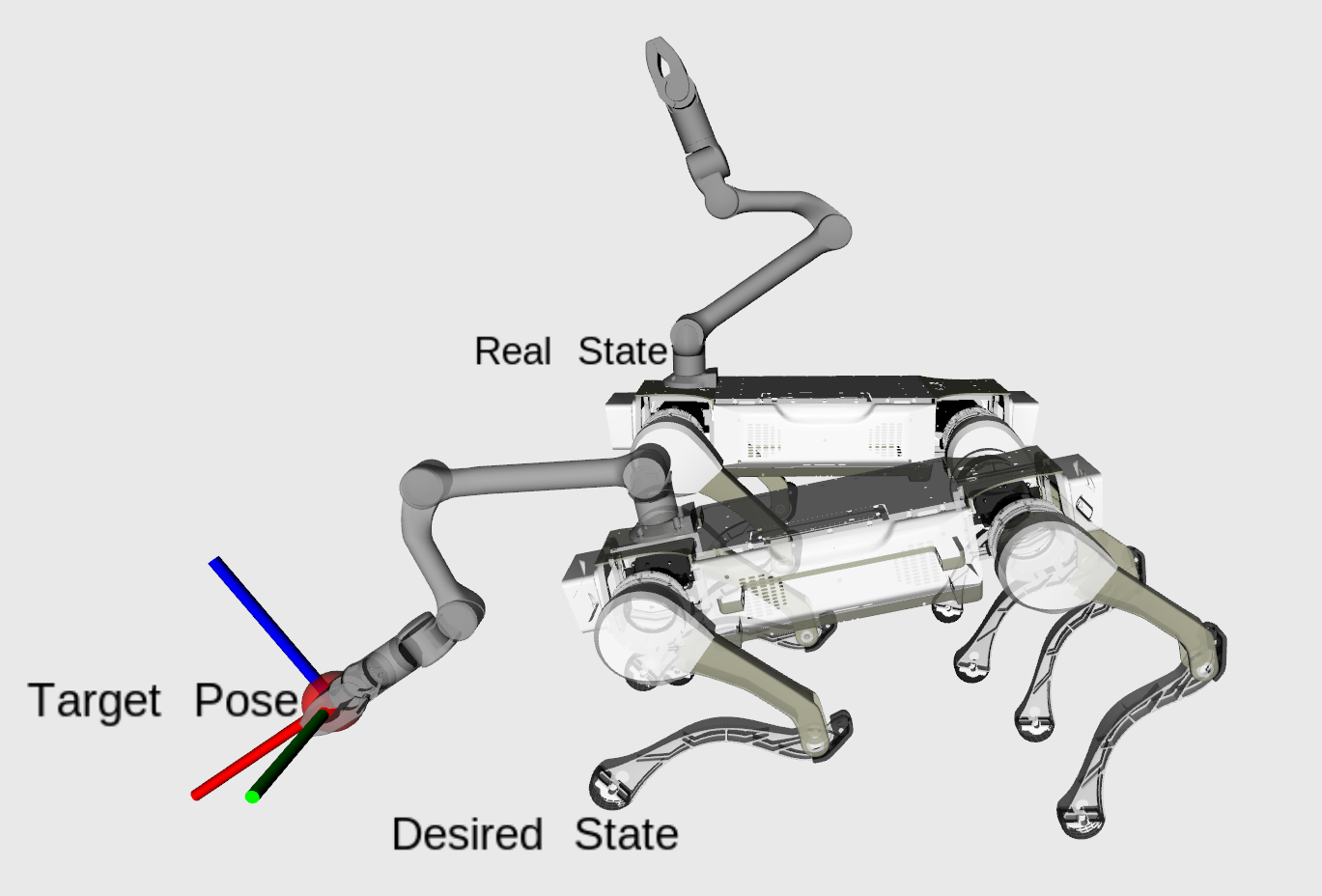}
            \includegraphics[height=2.76cm]{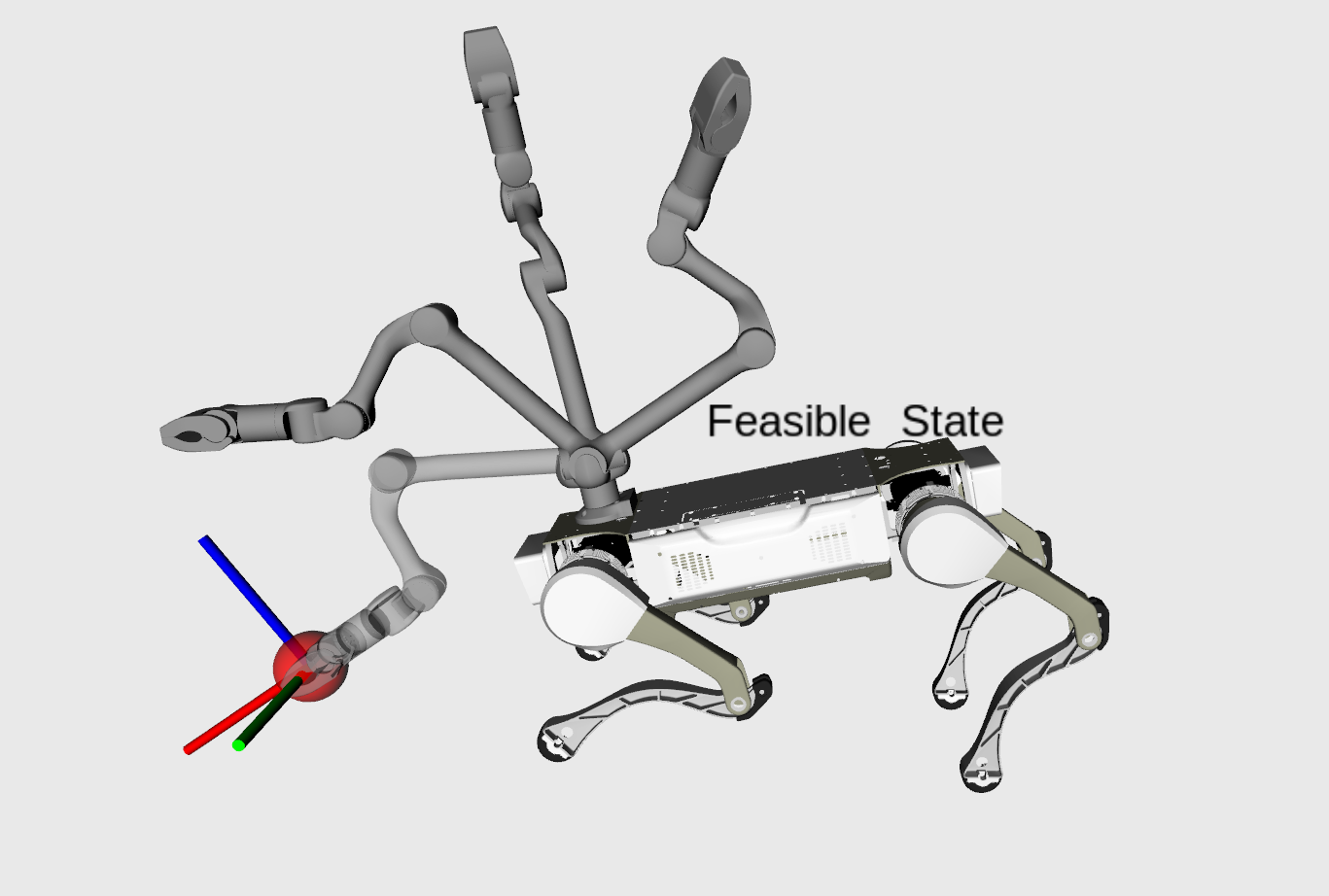}
        \end{minipage}
    }
    \vspace{-5mm}
    \caption{\textbf{Physical Feasibility Guided Reward Design}. In the first image, the robot is given with a target near the ground and the robot should bend its front legs, facilitating the manipulator's movement towards the target. The second image depicts the feasible state reward mechanism. When the body explores a feasible state, it is rewarded irrespective of the manipulator's position. This mechanism will facilitate the learning of whole-body coordination motion.}
    \label{Fig.pfg}
\end{figure}
To illustrate the difficulties in optimizing the whole-body control of legged manipulators, we consider a scenario where a target near the ground is given to the manipulator, and the quadruped robot is commanded to move at a certain speed, the robot should complete the movement while maintaining the tracking accuracy of the manipulator by actively bending the front leg, as shown in the first image of Fig. \ref{Fig.pfg}. However, in the initial phase of learning, the policy will prioritize learning to maintain body balance because the performance of both velocity tracking and end-effector pose control is immature. Although the posture of blending the front leg can expand the manipulator's manipulation space, it will significantly affect the optimization process of velocity tracking control, leading to the abandonment of such high-value body postures. Only after the policy has mastered the basic locomotion skills will it re-explore these synergistic postures driven by the manipulator's target-tracking rewards. This learning process seriously reduces learning efficiency.

To address this problem, 
we define a feasible state function\eqref{eq:feasible_states1} to map the relationship between the quadruped robot’s motion and the manipulator’s operational workspace. Specifically, given a target end-effector pose $^WT \in SE(3)$ in the world frame and the robot’s torso state $s_{torso}$, we consider $s_{torso}$ feasible if there exists a valid manipulator joint configuration $q_{arm}\in \Theta_{arm}$ such that $FK(s_{torso},q_{arm}) = {^WT}$, where $FK$ denotes the forward kinematics model of the legged manipulator and $\Theta_{arm}$ is the manipulator's joint configuration domain. When $FS(s_{torso}, {^WT}) =1$, the torso pose $s_{torso}$ is deemed to provide kinematic feasibility for the manipulator to eventually reach the target pose, even if not yet achieved physically, described in the second image of Fig. \ref{Fig.pfg}. By rewarding such feasible torso states during training, the algorithm guides RL policy toward discovering collaborative motor skills more efficiently in early exploration stages. We refer to this method as the physical feasibility-guided(PFG) reward design method.
\begin{align}
\boldsymbol{FS}(s_{torso}, ^WT) &= \begin{cases} 
1, & \text{if } \exists q_{arm}\in\Theta_{arm},
\\&\text{s.t. } FK(s_{torso},q_{arm})=^WT \\
0, & \text{otherwise}
\end{cases}\label{eq:feasible_states1}
\\&=\begin{cases} 
1, & \text{if } \exists q_{arm}\in\Theta_{arm},
\\&\text{s.t. } FK_{arm}(q_{arm})=^BT(s_{torso}) \\
0, & \text{otherwise}
\end{cases} \label{eq:feasible_states2}
\end{align}

To simplify implementation, we transform the target pose ${^WT}$ into the torso-dependent body frame ${^BT(s_{torso})}$\eqref{eq:feasible_states2}, eliminating the need for a full-body kinematic model and focusing solely on the manipulator’s kinematics. We employ a product of exponentials (POE)\cite{brockett2005robotic} formulation to model the manipulator’s generic kinematics and solve the inverse kinematics(IK) problem using a damped least-squares algorithm\cite{chan1988general,sekiguchi2021numerical,haviland2023manipulator}. This framework ensures compatibility with diverse legged manipulators. The IK algorithm is outlined in Algorithm \ref{alg1}, where $J$ is the Jacobian matrix, $^BT \in SE(3)$ is the end-effector pose under body frame, $\vec{e_t}$ the exponential coordinates of ${}^BT^{-1}_{t} {}^BT$, $D_t$ and $g_t$ are intermediate variables. The operator $\ominus$ denotes generalized pose subtraction in $SE(3)$. For two poses $T_{a}, T_{b} \in SE(3)$, $T_{a} \ominus T_{b}$ are defined as $T_{a} \ominus T_{b} = log(T_b^{-1} T_a) \in R^6$.
\renewcommand{\algorithmicrequire}{\textbf{Input:}}
\renewcommand{\algorithmicensure}{\textbf{Output:}}
\begin{algorithm}[!t]
\caption{Inverse Kinematic Algorithm}
\label{alg1}
\begin{algorithmic}[1]
  \REQUIRE  Expected pose $^BT_{t}$ at the time $t$; Joint position of the manipulator $q_{real}$; The solved joint position $^{ideal}q_{t-1}$ at the time $t-1$; Twist error weight $W_{e}$; 
  \ENSURE Expected joint position $^{ideal}q_{t}$; Does inverse kinematics have a solution $b_{t}$;
  \STATE {$^{ideal}q_{t} = ^{ideal}q_{t-1}$}
  \FOR{$step \leq 10$}
    \STATE {$^BT = ForwardKinematic(^{ideal}q_{t})$}
    \STATE {$\vec{e_t} = ^BT_{t} \ominus ^BT$}
    \STATE {$e_t = 0.5*\vec{e_t}^T W_e \vec{e_t}$}
    \IF {$e_t <= 0.001$}
       \RETURN {$^{ideal}q_{t}, b_{t} = true$}
    \ENDIF
    \IF {$^{ideal}q(i)$ reach joint limit}
      \STATE {$J.column(i) = 0$}
    \ENDIF
    \STATE $D_t = J^T W_{e} J + 0.5(\vec{e_t}^T W_e \vec{e_t} + \delta)I$
    \STATE $g_t = J^T W_{e} \vec{e_t}$
    \STATE $^{ideal}q_{t} = ^{ideal}q_{t} + D_t^{-1} g_t$
  \ENDFOR
  \RETURN {$^{ideal}q_{t}= q_{real}, b_{t} = false$}
\end{algorithmic}
\end{algorithm}

\subsection{Rewards} \label{reward_section}
\begin{table}[!t]
\caption{Detailed Reward Terms}
\label{reward_table}
\vspace{-5mm}
\begin{center}
\begin{tabular}{ccc}
\hline 
Reward Items & Definition & Weight\\
\hline
\rowcolor{lightgray!40}\multicolumn{3}{c}{Feasible State Reward $r_{feasible}$} \\
\textbf{kinematics} & $\max(e^{\sum{|^{ideal}q_{i}-q_i|}}, 0.2)$ & $0.16$ \\
\rowcolor{lightgray!40}\multicolumn{3}{c}{Task Rewards $r_{task}$} \\
linear velocity & $e^{\Vert v_{cmd} - v \Vert}$ & $0.5$ \\
angular velocity & $e^{(\omega_{cmd}-\omega)^2}$ & $0.3$ \\ 
pose tracking & $e^{\Vert T_{cmd} \ominus T \Vert}$ & $0.6$ \\
\rowcolor{lightgray!40}\multicolumn{3}{c}{Regularization Rewards(x20Z1) $r_{reg}$}\\
torque & $\sum \tau_i^2$ & $-1.2 \times10^{-5}$ \\
torque smooth & $ \sqrt{\sum\mathrm{\dot{\tau}}_i^2} $ & $-1.5 \times 10^{-6}$ \\
joint acceleration & $\sum{\ddot{q}_i^2}$ & $-1.0 \times 10^{-7}$ \\
joint limit& $\sum{(q_i<0.96q_{min})||(q_i>0.96q_{max})}$ & $-0.1$ \\
collision & $\sum c_i$ & $-0.4$ \\
clearance & $\sum d_i,i\in{1,2,3,4}$ & $3.0$ \\
lift time & $\sum t_i, i\in{1, 2, 3, 4}$ & $0.35$ \\
slip & $\sum s_i, i\in{1, 2, 3, 4}$ &  $-0.15$ \\
\hline 
\end{tabular}
\end{center}
\end{table}

The reward function comprises three components. The first is the feasible state reward $r_{feasible}$, which is awarded if and only if $FS(s_{torso}, ^WT)=1$. This term is always positive and serves two purposes: (1) guiding reinforcement learning exploration to facilitate the learning of whole-body coordination, and (2) leveraging inverse kinematics-derived desired joint states to enhance the manipulator’s end-effector pose tracking accuracy. The second component, the task goal reward $r_{task}$, includes three sub-rewards: $r_{\text{linear velocity}}$ and $r_{\text{angular velocity}}$, which penalize deviations from the quadruped robot’s desired linear and angular velocities, and $r_{\text{pose tracking}}$ which quantifies the end-effector’s pose tracking error. The final component is the regularization reward $r_{reg}$, designed to refine robot behavior. For example, collision reward penalties to prevent undesired interactions between the body and the arm and torque reward penalties to minimize energy consumption. Regularization reward is critical for generating smooth, stable motions and is needed to be tuned for different legged manipulators. For the X20-Z1 legged manipulator, Table \ref{reward_table} details the definitions and weighting coefficients $w_i$ of each reward term. In TABLE \ref{reward_table}, $v$ represents linear velocity, $\omega$ represents angular velocity, $c$ the number of collision points excluding the four feet, $d_i$ the distance from the ground when the $i$th leg is lifted, $t_i$ the time at which the $i$th leg is lifted, $s_i$ the tangential velocity when the $i$th leg touches the ground, $\tau$ represents the joint moment, $T$ is the pose of end-effector, $q$ are the joint positions, and $\ddot{q}$ are joint accelerations. The total reward is computed as: $r = \sum_i w_i r_i$.

\subsection{Command Planner} \label{cmd_planner_section}
\begin{figure}[!t]
    \centering
    \subfigure{
        \centering
        \begin{minipage}{\linewidth}
            \includegraphics[height=3.7cm]{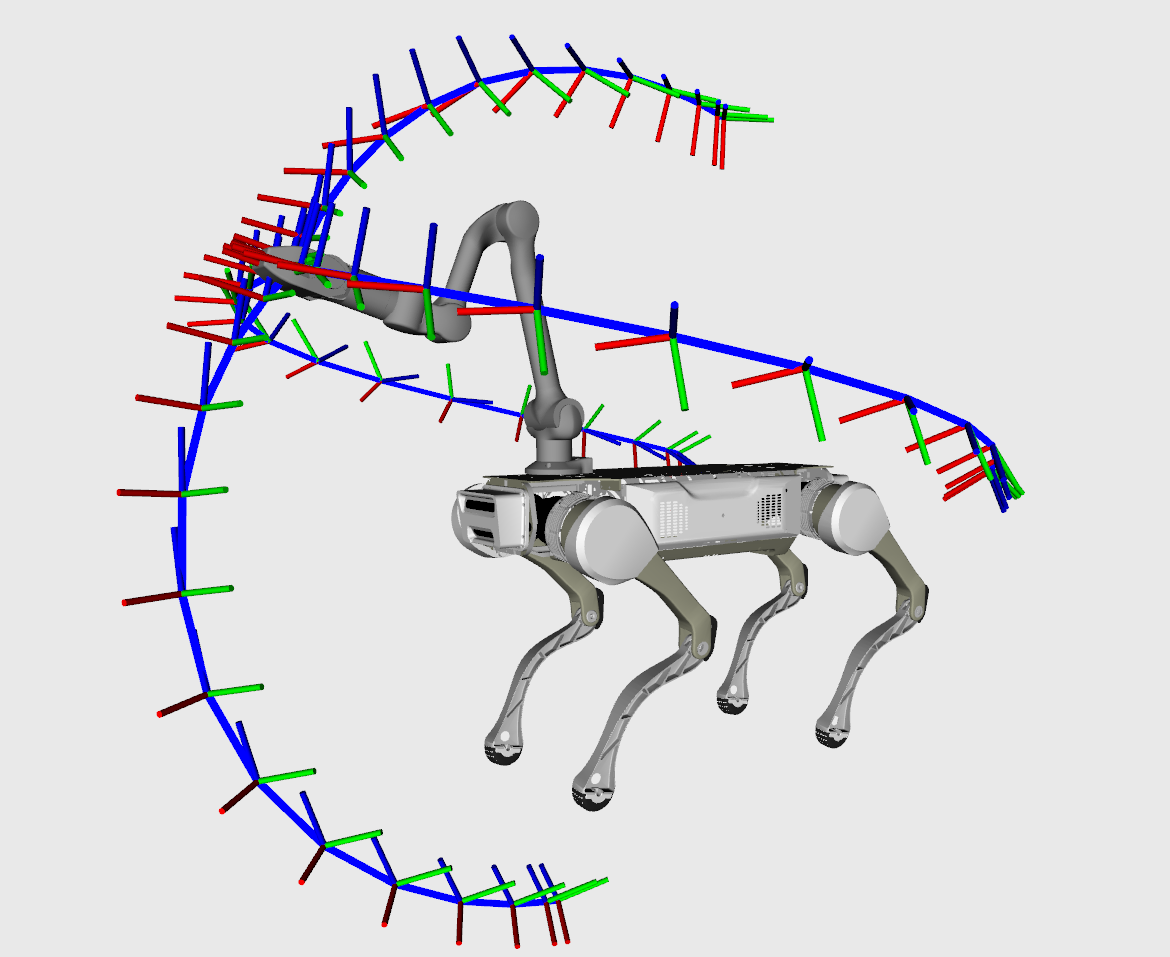}
            \includegraphics[height=3.7cm]{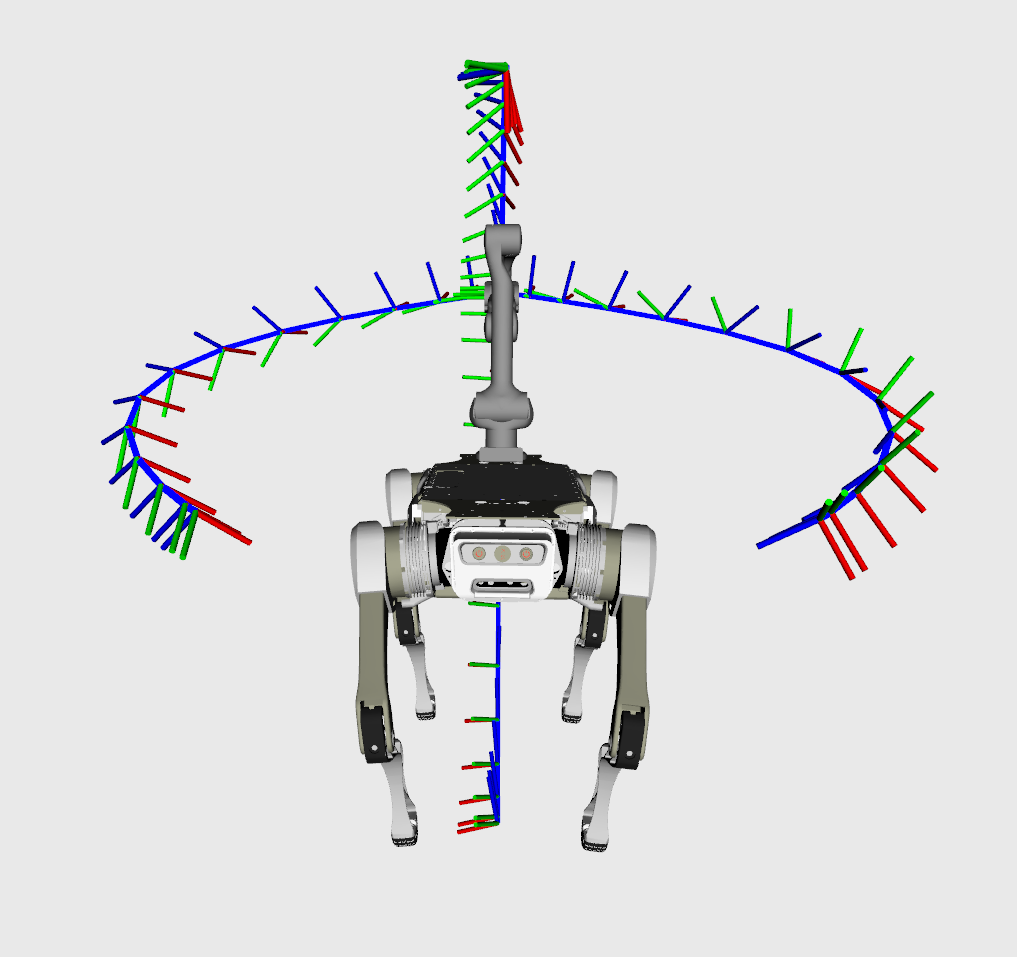}
        \end{minipage}
    }
    \vspace{-5mm}
    \caption{\textbf{Command Planner}. These two pictures show four smooth reference trajectories generated by the command planner module. Smooth reference trajectories mitigate the learning complexity in the manipulator tracking task while simultaneously enhancing the smoothness of control outputs.}
    \label{Fig.cmd_planner}
    \vspace{-5mm}
\end{figure}
To enhance the efficiency of learning inverse kinematics, 
we propose a body-projected coordinate system-based method for reference trajectory generation. The $\{Proj\}$ coordinate system is constructed by orthogonally projecting the body coordinate system onto the terrain surface. Its origin is always fixed at the ground surface, with the yaw axis is aligned to the body's orientation, while the pitch and roll axes remain horizontal. At the beginning of each training cycle, the initial pose of the manipulator's end-effector is recorded in the $\{Proj\}$ frame as $^{Proj}T_{init}\in SE(3)$ and a target pose $^{Proj}T_{end} \in SE(3)$ is sampled to generate the reference trajectory. The position part of $^{Proj}T_{end}$ is sampled randomly in a $1.0$m sphere centred on the body frame, and the rotational part is sampled uniformly in SO(3) space. As the robot moves, the $\{Proj\}$ system translates synchronously with the robot, ensuring the reference trajectory is always within a reasonable range. Trajectory interpolation is achieved via cubic polynomial algorithms \eqref{eq:reference_trajectory}, producing smooth intermediate waypoints. In \eqref{eq:reference_trajectory}, $\oplus$ denotes the bit-pose synthesis operation on $SE(3)$ manifolds, $t$ represents the time, $t_{total}$ the trajectory time, and $T_{ref}$ is the reference trajectory. Fig.\ref{Fig.cmd_planner} shows four reference trajectories generated by the command planner.
\begin{align}
\label{eq:reference_trajectory}
T_{\mathrm{ref}}(t) &= T_{\mathrm{init}} \oplus \exp\left[ 
\frac{3\Delta T}{t_{\mathrm{total}}^2} t^2 - \frac{2\Delta T}{t_{\mathrm{total}}^3} t^3 
\right] \\ \Delta T &= T_{\mathrm{end}} \ominus T_{\mathrm{init}}
\end{align}

\section{EXPERIMENTS} \label{experiments_section}
We use the Raisim\cite{raisim} physics engine to concurrently simulate 2048 parallel training environments. Training was carried out on a system with an AMD 7950X processor and a 3090 GPU, taking approximately 9 hours to acquiring a policy. Stable policies were successfully learned for both Aliengo and X20 quadruped robots equipped with the Unitree Z1 manipulator in simulation. The policy was deployed zero-shot on an X20-Z1 robot in real-world testing.

\subsection{Ablation Studies}
\begin{figure}[!t]
    \centering
    \includegraphics[width=\linewidth]{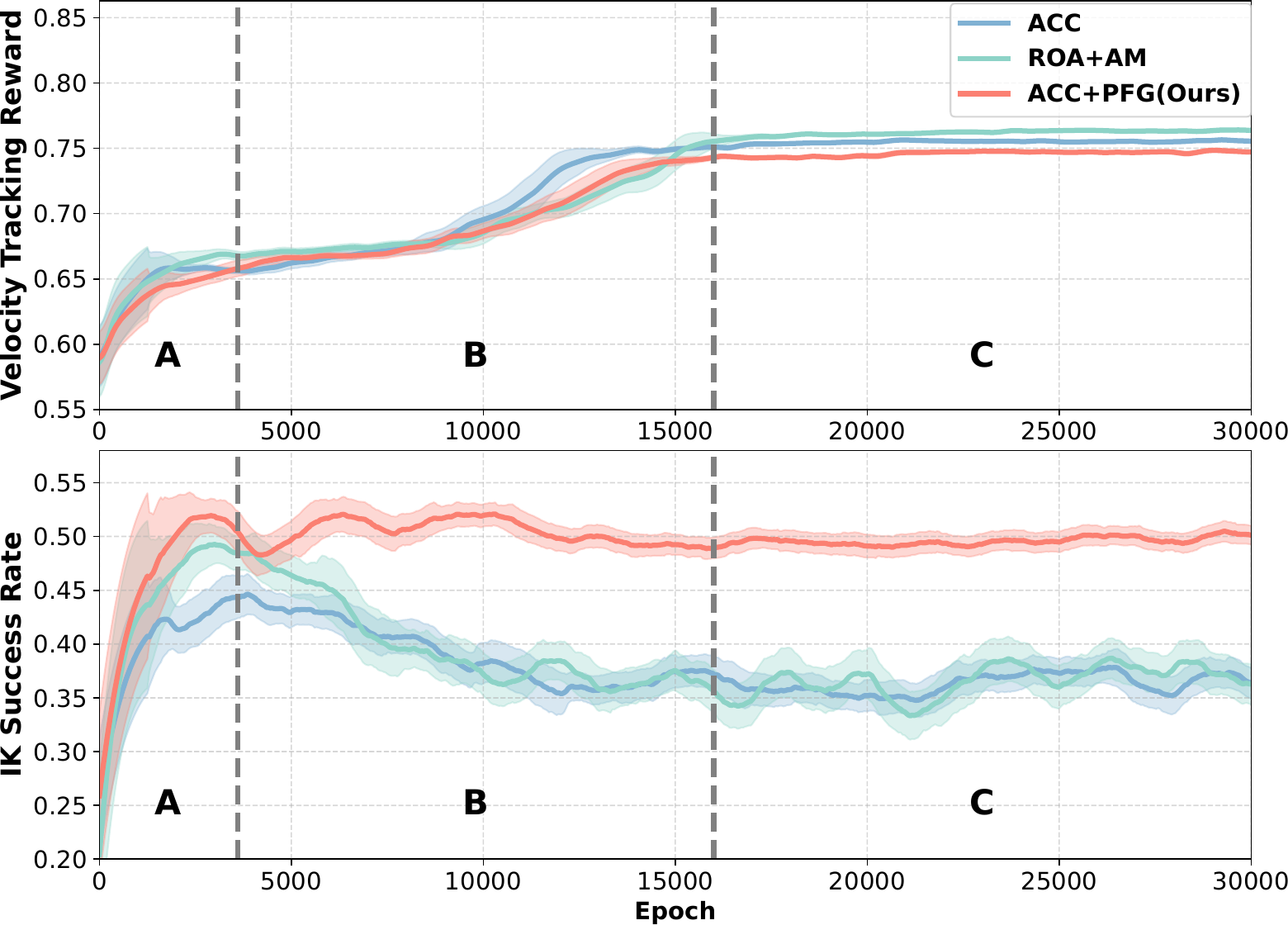}
    \vspace{-5mm}
    \caption{\textbf{Ablation Studies}. PFG maintains the robot's cooperative behavior during locomotion optimization, ensuring the training avoids local optima.}
    \label{Fig.ablation}
\end{figure}
To evaluate the effectiveness of the PFG method in preventing the algorithm optimization process from converging to local minima, two key metrics were designed: 1. \textbf{Inverse Kinematics(IK) Solution Rate}: The command planner samples the end-effector's target in real-time and calculates the proportion of successful inverse kinematics solutions during the training process. This metric reflects legged manipulator's ability to achieve cooperative motion. A higher IK solution rate indicates that the robot can more effectively expand its manipulation space by adjusting its body posture, thereby assisting the manipulator in completing pose tracking tasks. 2. \textbf{Velocity tracking reward}: This metric quantifies the quadrupedal robot's velocity tracking accuracy and evaluates the impact of the cooperative manipulation behaviors of the legged manipulator on the quadruped's velocity control. The following three sets of comparison experiments were designed:
\begin{itemize}
    \item Asymmetric Actor Critic(AAC): A baseline approach without integrating the PFG reward.
    \item AAC+PFG: The AAC framework integrated with the PFG reward to validate its optimization benefits.
    \item Regularized Online Adaptation with Advantage Mixing (ROA+AM)\cite{fu2023deep}: A comparative approach to assess the performance of AAC+PFG.
\end{itemize}

As illustrated in Fig. \ref{Fig.ablation}, the training process can be divided into three distinct phases. Phase A: In this initial phase, the policy focuses on developing the ability to maintain the balance. Velocity tracking control and end-effector pose control exhibit relatively low performance. However, the IK solution rate shows a consistent upward trend, indicating that the RL algorithm is gradually learning body-arm coordination through exploration. In this phase, the IK solution rate for the ACC+PFG method increases to $51\%$, compared to only $43\%$ for the ACC method without the PFG reward. This suggests that the PFG reward enhances the policy's ability to explore whole-body cooperative motion. Similarly, the ROA+AM method also supports the learning of whole-body cooperative behaviors, and its effectiveness is comparable to that of the ACC+PFG method.

Phase B: In this phase, there is a notable increase in velocity tracking rewards. However, the IK solution rates for both the ACC and ROA+AM methods drop to $37\%$, indicating that the robot prioritizes achieving better locomotion rewards over maintaining the cooperative behaviors necessary for end-effector tracking accuracy. In contrast, the PFG reward enables IK solution rate remain stable, enabling the robot to preserve whole-body cooperative behaviors and avoid convergence to local minima during optimization. 

Phase C: In the final stage, the policy undergoes fine-tuning guided by regularization rewards to ensure compliance with multiple constraints, such as energy efficiency. During this phase, the IK solution rate and velocoty tracking rewards stabilize across all methods. Experimental results demonstrate that the PFG reward enables the policy to extensively explore diverse whole-body cooperative motions of the legged manipulator, significantly expanding the workspace of the legged manipulator by $34\%$ (based on the IK solution rate), while maintaining velocity tracking rewrad degradation at $\le3\%$, compared to the ACC and ROA+AM methods.

\subsection{Performance Validations}
\begin{table}[!t]
\caption{\textbf{Performance Validations}}
\label{Tab.performance_validation}
\vspace{-5mm}
\begin{center}
\begin{tabular}{ccccc}
\hline 
~& PE(m)$\downarrow$ & RE(rad)$\downarrow$ & LVTE(m/s)$\downarrow$ & AVTE(rad/s)$\downarrow$ \\ 
\hline
AAC+PFG & $\textbf{0.087}$ & $\textbf{0.18}$ & $0.43\pm0.11$ & $\textbf{0.12}\pm\textbf{0.04}$ \\
ROA+AM & $0.144$ & $0.33$ & $\textbf{0.35}\pm\textbf{0.12}$ & $0.16\pm0.10$ \\
AAC & $0.131$ & $0.21$ & $0.45\pm0.08$ & $0.15\pm0.06$ \\
ROA & $0.144$ & $0.30$ & $0.42\pm0.15$ & $0.16\pm0.15$  \\ 
\hline 
\end{tabular}
\end{center}
\end{table}
\begin{figure}[!t]
    \centering
    \includegraphics[width=\linewidth]{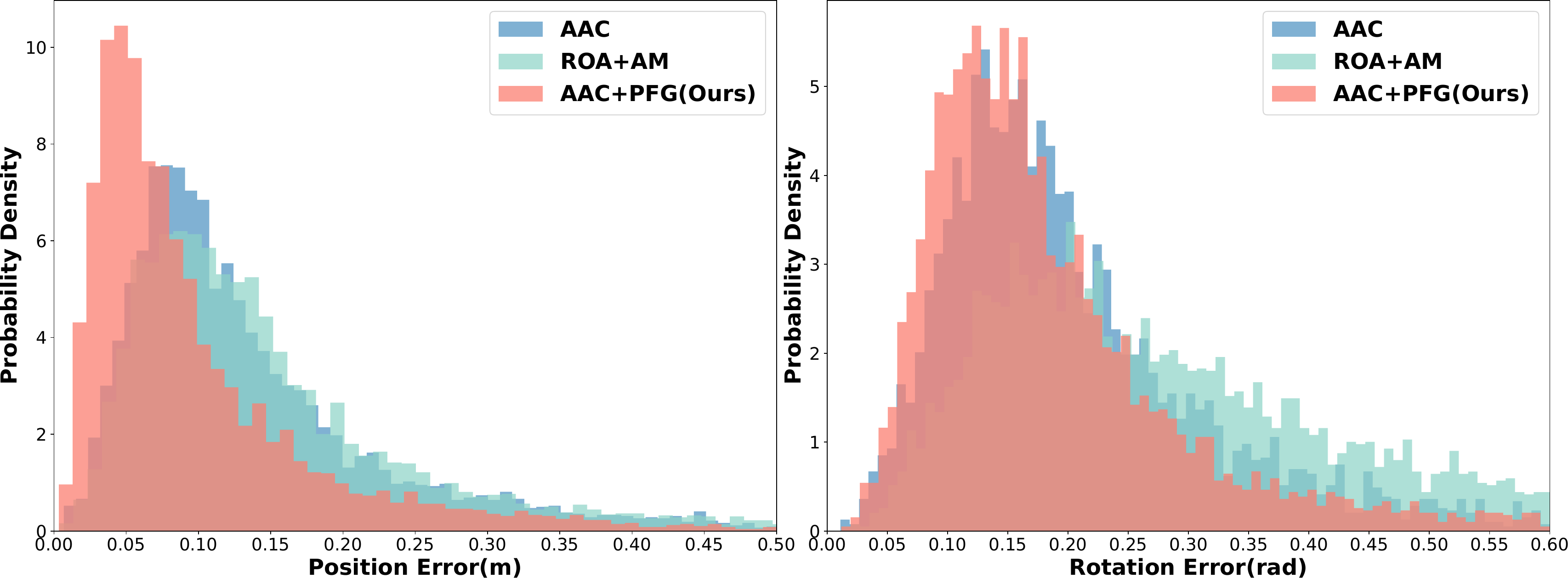}
    \vspace{-5mm}
    \caption{\textbf{Pose Tracking Accuracy}}
    \label{Fig.accuracy}
    \vspace{-5mm}
\end{figure}
To evaluate the model's ability to independently control the body and the manipulator, two experimental metrics were employed for evaluation: 1.\textbf{Target Pose Tracking Accuracy}: In the simulation, 5000 target poses were randomly generated based on the forward kinematics of the manipulator. The positional error (PE), evaluated using Euclidean distance, and the rotational error (RE), evaluated using the SO(3) geodesic distance, were statistically calculated for the model's control performance. The 60th percentile was used as the accuracy evaluation criterion. 2.\textbf{Velocity Tracking Accuracy}: For the quadruped robot, $1000$ velocity commands were randomly sampled within the ranges of [-1.3 m/s, 1.3 m/s] for linear velocity and [-1.3 rad/s, 1.3 rad/s] for angular velocity, while the manipulator was kept stationary. The linear velocity tracking error (LVTE) and angular velocity tracking error (AVTE) were computed to access model performance.

As illustrated in Table \ref{Tab.performance_validation}, a comparison of the four methods(ACC+PFG, ACC, ROA and ROA+AM) reveals the following results: The PFG reward enhances the manipulator's position tracking accuracy, achieving a position error of $\le8.7$ cm and a rotation error of $\le0.18$ rad in $60\%$ of the manipulator's workspace. 
In contrast, the Advantage-Mixing(AM) approach effectively improves body linear velocity tracking accuracy but does not contribute to reducing the end-effector's tracking error. The tracking accuracies distribution of the ACC, ROA+AM, and ACC+PFG methods are illustrated in Fig.\ref{Fig.accuracy}, demonstrating that the end-effector control accuracy of the RL method is not uniformly distributed across the manipulator workspace. Our experiments indicate that the tracking accuracy is highest in the anterior region of the quadruped's body, where the reference trajectory is predominantly concentrated, while performance degrades in the posterior-inferior region. 
Although adjusting the body position through velocity commands can compensate by placing the target in the high-precision region, the inability to achieve consistent accuracy across the entire manipulator workspace remains a limitation of RL in the manipulation applications of legged manipulators.

\subsection{Hardware Validations}
\begin{figure}
    \centering
    \vspace{10pt}
    \subfigure{
    \includegraphics[width=0.97\linewidth]{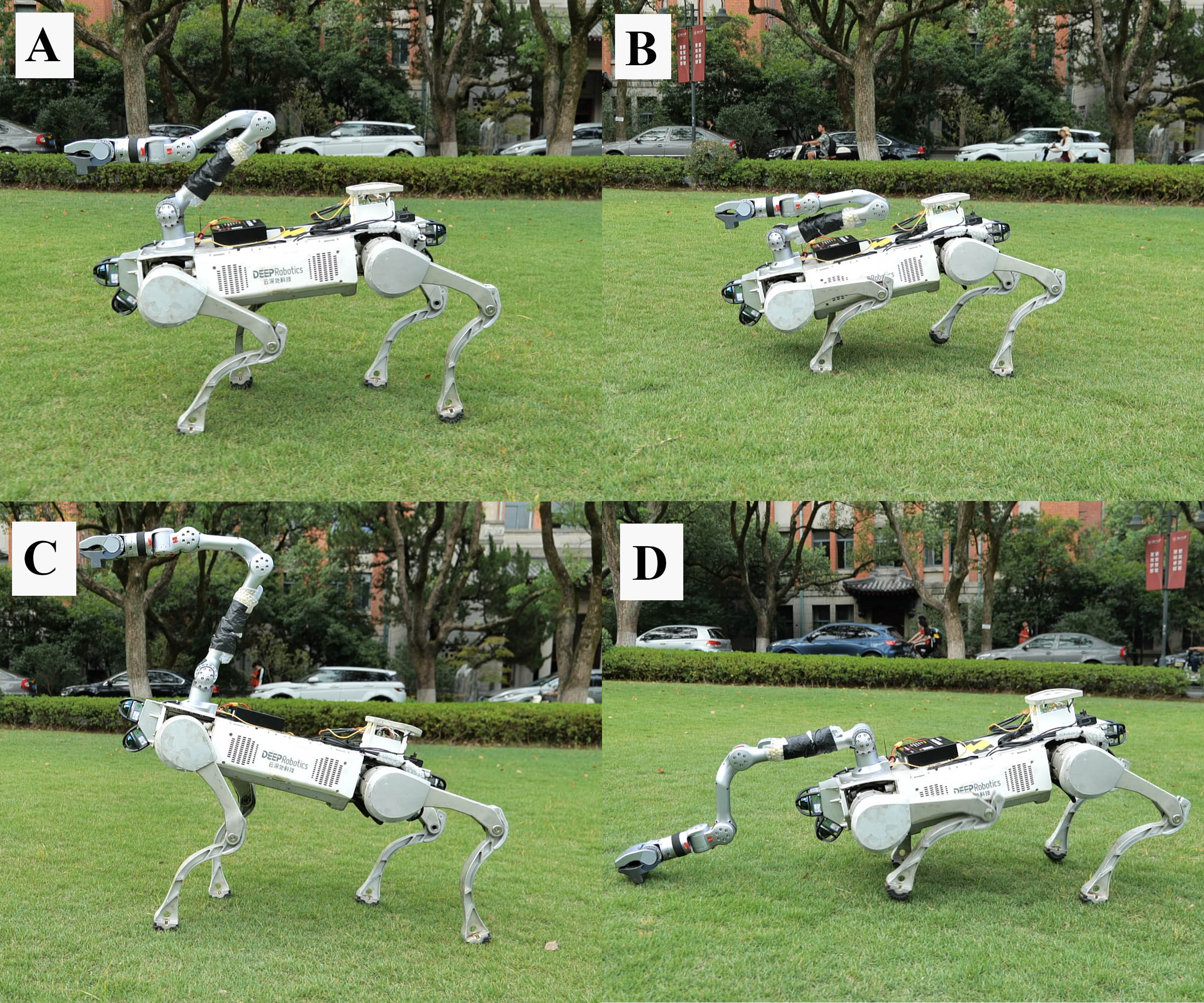}
    }
    \\ 
    \centering
    \subfigure{
      \includegraphics[width=0.97\linewidth]{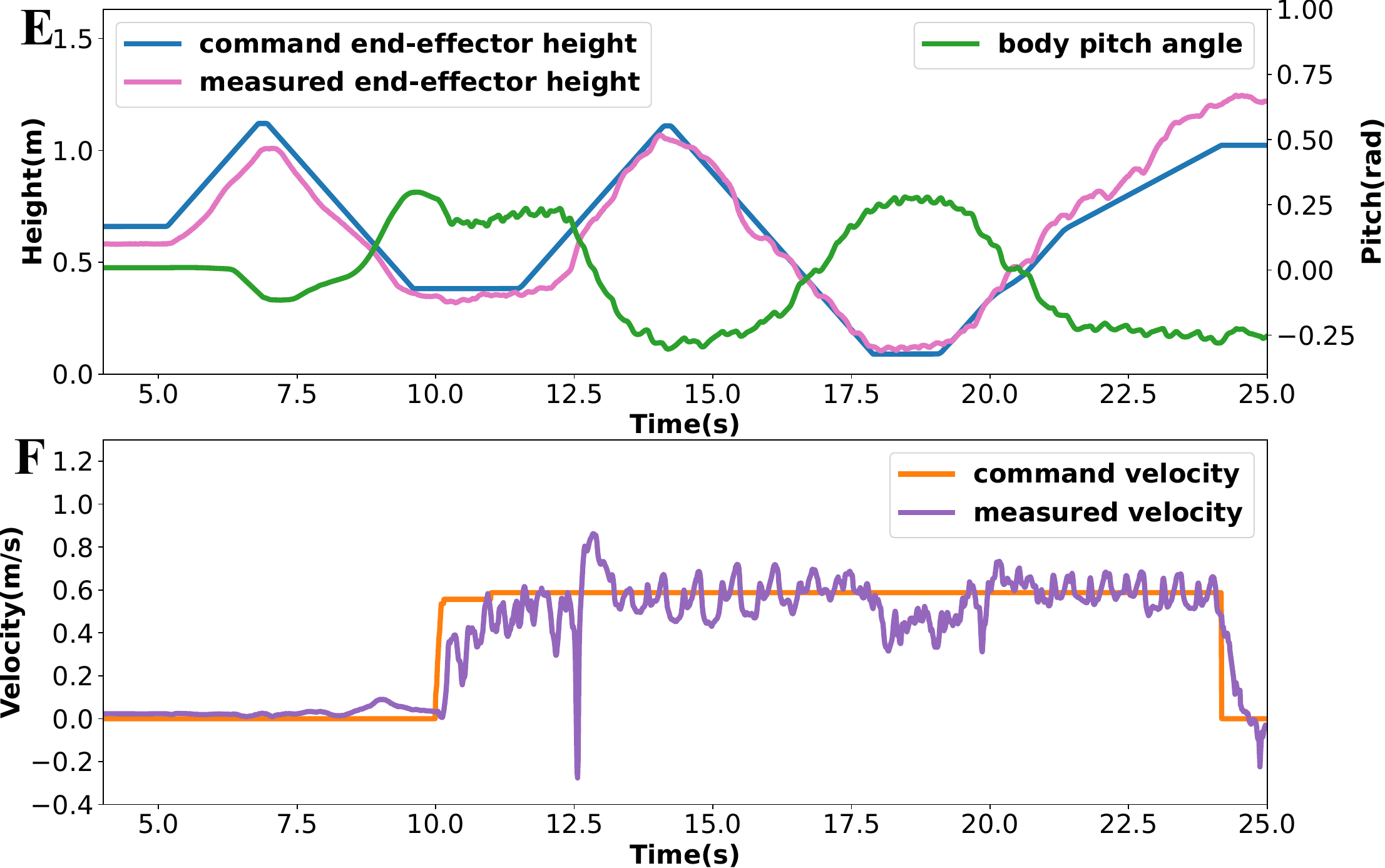}
    }
    \vspace{-5mm}
    \caption{\textbf{Smoothly coordinated whole-body motions in various postures during both static and dynamically configurations}. (A-D) illustrates various reaching motions with coordinated whole-body movements demonstrated by the robot during experiments. (E) depicts trajectory tracking performance of the robotic arm with naturally adaptive body pitch motion. (F) illustrates the velocity tracking performance of the body.}
    \label{fig: coordination motion}
    \vspace{-5mm}
\end{figure}

We conducted experiments with a real robot to evaluate the performance of our proposed method. The robot was commanded to move forward from a standing position while its manipulator executed vertical movements across a wide height range. Fig. \ref{fig: coordination motion}(A-D) shows the robot's cooperative motions during the experiment. Fig. \ref{fig: coordination motion}(E) depicts the height tracking curves of the manipulator's end-effector, along with the corresponding data on the robot's body posture. We observe that as the end-effector's height increases, the quadruped robot reduces its pitch angle, resulting in an upward-tilted posture. Conversely, when this height decreases, the pitch angle increases correspondingly, causing the body to adopt a downward-tilted posture. This coordinated behavior ensures that the manipulator maintains a high level of tracking performance during large-range movements. Although the manipulator's motion disrupts the body, the controller effectively compensates for this disturbance, maintaining stable body velocity, as demonstrated in Fig. \ref{fig: coordination motion}(F). Additionally, as illustrated in Fig.\ref{fig:whole-body skills}, our approach has been successfully applied to  a variety of tasks, including mimicking human gymnastic movements, waving 4-meter-long ribbons, performing v-log film task, picking up objects from the ground, and moving loaded trolleys.

\section{DISCUSSION AND FUTURE WORKS}
We develop a unified RL framework that integrates explicit kinematics models, enabling efficient whole-body skills of a quadruped with a manipulator, which effectively incorporates physics-based information in the form of kinematics and is promising for future extensions. Future work will extend physics-informed learning and integrate more detailed physics models into learning structures, ensuring neural network outputs adhere to physical laws. This new capability will autonomously produce highly dynamic and coherent whole-body motor skills, and allow us to develop more sophisticated high-level planners to navigate complex environments for a wide range of real-world operations. 

\addtolength{\textheight}{-9cm}  
\clearpage

\bibliographystyle{IEEEtran.bst}
\bibliography{IEEEabrv, references}
\end{document}